\pdfoutput=1
\documentclass[11pt]{article}

\usepackage[]{EMNLP2023}

\usepackage{times}
\usepackage{latexsym}

\usepackage[T1]{fontenc}
\usepackage[utf8]{inputenc}
\usepackage{graphicx} 
\usepackage{multirow}
\usepackage{caption}
\usepackage{subcaption}

\usepackage{booktabs}       % professional-quality tables

\usepackage{microtype}

\usepackage{inconsolata}

\title{Psychological Metrics for Dialog System Evaluation}

\author{Salvatore Giorgi${^\dagger}$, Shreya Havaldar${^\dagger}$, Farhan Ahmed${^\ddagger}$, Zuhaib Akhtar${^\diamond}$, \\  {\bf Shalaka Vaidya${^\diamond}$, Gary Pan${^\diamond}$, Lyle H. Ungar${^\dagger}$, H. Andrew Schwartz${^\ddagger}$, Jo\~{a}o Sedoc${^\diamond}$} \\
        ${^\dagger}$University of Pennsylvania, ${^\ddagger}$Stony Brook University, ${^\diamond}$New York University \\
        \texttt{ sal.giorgi@gmail.com, jsedoc@stern.nyu.edu}}
\begin{document}
\maketitle

\begin{abstract}
We present metrics for evaluating dialog systems through a psychologically-grounded ``human'' lens in which conversational agents express a diversity of both \textit{states} (e.g., emotion) and \textit{traits} (e.g., personality), just as people do. 
We present five interpretable metrics from established psychology that are fundamental to human communication and relationships: emotional entropy, linguistic style and emotion matching, agreeableness, and empathy. 
These metrics can be applied (1) across dialogs and (2) on turns within dialogs.
The psychological metrics are compared against seven state-of-the-art traditional metrics (e.g., BARTScore and BLEURT) on seven standard dialog system data sets. 
We also introduce a novel data set, the {\it Three Bot Dialog Evaluation Corpus}, which consists of annotated conversations from ChatGPT, GPT-3, and BlenderBot.
We demonstrate that our proposed metrics offer novel information; they are uncorrelated with traditional metrics, can be used to meaningfully compare dialog systems, and lead to increased accuracy (beyond existing traditional metrics) in predicting crowd-sourced dialog judgements. The interpretability and unique signal of our psychological metrics make them a valuable tool for evaluating and improving dialog systems.

% We present metrics for evaluating dialog systems through a psychologically-grounded ``human'' lens in which conversational agents express a diversity of both \textit{states} (short-term factors like emotions) and \textit{traits} (longer-term factors like personality) just as people do. 
% These interpretable metrics consist of five measures from established psychology and are fundamental to human communication and relationships.
% The metrics that can be applied both across dialogs and on turns within dialogs: emotional entropy, linguistic style and emotion matching, agreeableness, and empathy. 
% We compare these human metrics against seven state-of-the-art traditional metrics (e.g., BARTScore and BLEURT) on nine standard dialog system data sets. 
% We also introduce a novel data set, the {\it Three Bot Dialog Evaluation Corpus}, which consists of annotated conversations from ChatGPT, GPT-3, and BlenderBot.
% We demonstrate that the proposed human metrics offer novel information, are uncorrelated with traditional metrics and lead to increased accuracy beyond existing traditional metrics for predicting crowd-sourced dialog judgements.
% In the absence of human judgements, the interpretable human metrics offer additional insights, allowing for comparisons across dialog systems, which is not always possible with traditional metrics.
% The interpretability and unique signal of our proposed human-centered metrics make them a valuable tool for evaluating and improving dialog systems.

\end{abstract}

%$\keywords{First keyword \and Second keyword \and More}

\section{Introduction}
\label{sec:intro}

%main motivation

Metrics that capture human-like attributes of dialog agents can help inform dialog agents that better converse and connect with users. Evaluating dialog agents from a ``human'' lens can help identify areas where current systems fall short. For example, dialog agents that are not empathetic or fail to match the linguistic style of a user will struggle to succeed as therapy bots or teaching aids. Large language models today are becoming increasingly conversant, and so we require efficient metrics to properly evaluate these conversations from a psychological perspective.

\begin{figure}[!tb]
\centering
\includegraphics[width=\linewidth]{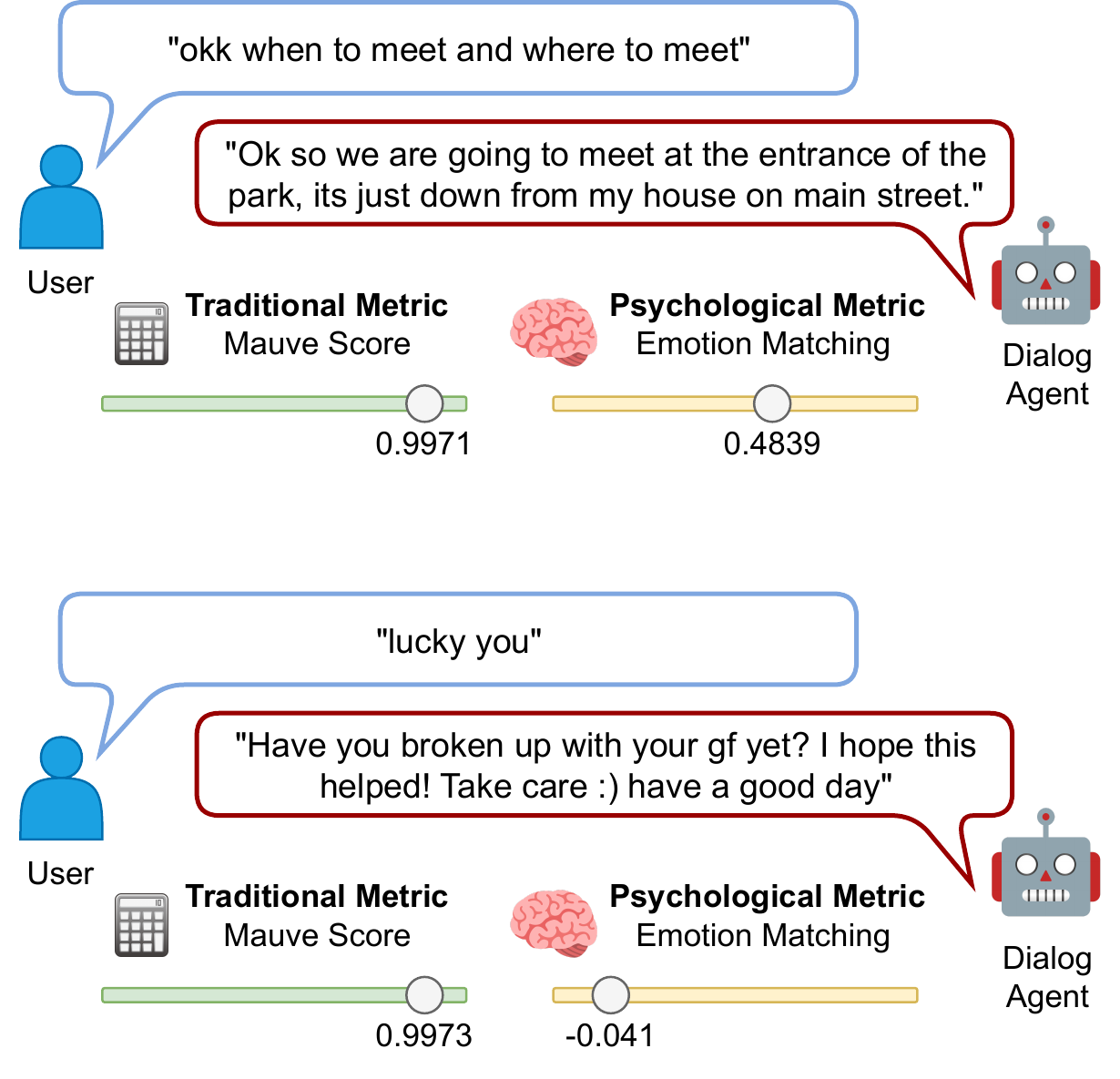}
\caption{Mauve score (traditional metric) and emotion matching (psychological metric) to evaluate two conversation snippets (turns). Humans rated the top response as highly appropriate and the bottom response as inappropriate. The dialog agent's response in both conversations receives a near-identical Mauve score but very different Emotion Matching scores. The disparity  between traditional metrics and human judgements highlights the need for psychologically-grounded metrics to evaluate dialog agents.}
\label{fig:metric-comparison}
\end{figure}

Open-domain dialog systems are typically evaluated using traditional automatic metrics (e.g., BLEU, METEOR, and ROUGE) or human judgements. However, both of these techniques have several drawbacks~\cite{zhang2021auteval}. 
Traditional automatic metrics aim to capture Gricean maxims (Quantity, Quality, Relation, and Manner); however, these maxims fail to capture the psychological aspects of a conversation. Additionally, these traditional metrics often rely heavily on overlap (e.g., word overlap for BLEU and semantic similarity for BERTScore) and fail to capture the diversity of dialog systems~\cite{liu2016not}. This limitation typically results in small associations with human judgements~\cite{liu2016not,deriu2021survey}. 
On the other hand, human judgements are expensive to scale and lack standardization~\cite{sedoc2019chateval,howcroft-etal-2020-twenty,smith-etal-2022-human}.
Automatic metrics that capture human-like dialog agent attributes could drive scalable and functional dialog system improvements.

%what we do broadly
In this work, we propose a set of psychologically-grounded metrics for evaluating open-domain dialog systems from a human lens, taking queues from \citet{giorgi2021characterizing}, which characterizes Twitter spambots through a number of human states and traits. 
We additionally propose three general classes of psychologically-grounded measures to characterize our metrics: (1) \textit{states} (changing within a dialog, such as emotion), (2) \textit{traits} (slower to change, such as personality), and (3) linguistic matching (i.e., how well chatbots match the linguistic cues of the other entity in the conversation). 

We also introduce the Three Bot Dialog Evaluation Corpus, a benchmark dialog data set of conversations with ChatGPT, GPT-3, and BlenderBot annotated at both the turn- and dialog-level. 
To highlight the usability of our proposed psychological metrics, we systematically compare them against a set of seven traditional metrics on the Three Bot Dialog Evaluation Corpus, as well as seven additional publicly available data sets. 
Finally, we compare dialog systems \emph{without human judgements} using the traditional and psychological metrics alone and show that our psychological metrics give functional and interpretable insights into these systems, while traditional metrics fall short.

\paragraph{Contributions} Our contributions include: 
\begin{itemize}
\item Proposing three classes of psychologically grounded metrics with five specific metric instances within these classes
\item Releasing a new data set of conversations from state-of-the-art dialog systems (ChatGPT, GPT-3, and BlenderBot) with turn- and dialog-level annotations
\item Systematically evaluating our psychological metrics against seven existing metrics across seven data sets 
\end{itemize}
We show that (a) psychological metrics are uncorrelated with traditional metrics and (b) using psychological metrics in conjunction with traditional metrics leads to increased accuracy in predicting crowd-sourced dialog system judgements.
Together, these results show that our psychological metrics can be used in tandem with existing metrics to further characterize and improve dialog systems.

\begin{figure*}[ht]
    \centering
    \includegraphics[width=\textwidth]{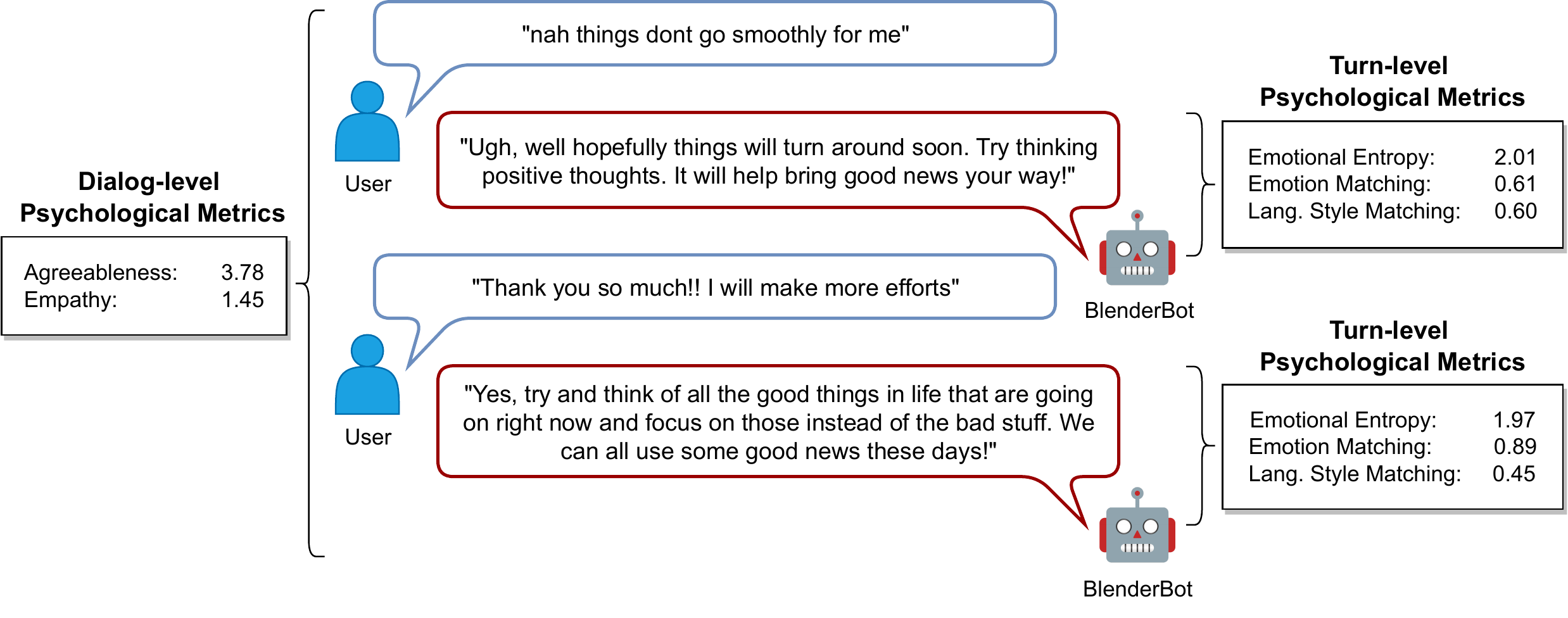}
    \caption{A sample from a dialog with BlenderBot, highlighting the hierarchical structure of dialogs (i.e., turns within dialogs). Here we see that turn-level metrics are calculated for each turn (as expected), while dialog-level psychological metrics are calculated across the entire conversation. This is typically not the case for traditional metrics, which are usually averaged across turns up to the dialog level.}
    \label{fig:enter-label}
\end{figure*}

\section{Related Work} 
\label{sec:related work}
There is a growing set of methods to embed language processing within human contexts~\cite{volkova2013exploring,hovy2015demographic,lynn-etal-2019-tweet}.
Most of such work has focused on modeling rather than on evaluation. For example, \emph{creating} agents with human-like traits such as empathy \cite{rashkin2019towards,omitaomu2022empathic}, trust \cite{novick2018inducing}, emotion \cite{zhouwang2018mojitalk,huber2018emotional}, and personalization and personas~\cite{li2016persona,zhang2018personalizing,mazare2018training,roller2021recipes}. 
In contrast, few have attempted to \emph{evaluate} dialog agents using human-like metrics. % having been proposed.
These few include \citet{adiwardana2020towards}, who proposed a metric that jointly measures ``making sense'' and being specific, both basic and important attributes of conversations. 
\cite{ghazarian-etal-2022-wrong} proposed a sentiment-based approach, which generalized to both spoken and written domains.
More directly, some have quantified ``humanness'' subjectively through crowd-sourcing: ``Which speaker sounds more human?'' ~\cite{li2019acute,roller2021recipes,deriu-etal-2020-spot}.

A parallel line of work seeks to improve language models by making them more human-aligned. \citet{santurkar2023opinions} evaluates whose opinions language models reflect via public opinion polls and \citet{Binz_2023} assesses whether language models reflect the cognitive ability of humans. \citet{glaese2022improving} establishes rules to make dialog agents more helpful and harmless. Additional work on assessing the alignment of agents \cite{askell2021general, ouyang2022training, bai2022constitutional} focuses on measuring and minimizing the attributes of agents that make them bad conversationalists (hate speech, toxicity, controversy, etc). 

Our work takes a different approach toward evaluating ``human-like'' dialog. 
We propose three classes of psychologically-grounded measures which can be used to evaluate dialog systems. These metrics additionally seek to measure and exemplify the attributes of dialog agents that make them \emph{good} conversationalists.
We see this as a step toward answering the call for a human-like open-domain system~\cite{adiwardana2020towards}, and for integrating current steps toward this.\footnote{For example, \citet{roller2021recipes} propose evaluating both ``engaging talking points'' and ``consistent persona'' which are captured within our proposed metrics via state and trait metrics, respectively, where consistency can be measured across multiple dialogs. }
Despite the general applicability of our proposed metrics, we note that a number of dialog systems are task or goal-oriented, such as question/answer systems~\cite{chen2017reading} or systems designed for highly specific tasks such as trip planning~\cite{el2017frames} and customer service~\cite{cui2017superagent}. Such systems may be considered outside of the scope of our formulation in that scheduling a trip is fundamentally different from, for example, talking to a conversational chatbot about COVID-19 vaccines, which may need additional social and cultural context. 

\section{Classes of Human-like Measures}
%Under the goal of a human-like open-domain dialog system, 
We propose two classes of measures: (1) states and traits and (2) linguistic matching, rooted in fundamental psychological measurements of humans and their social relationships and interactions (i.e., linguistic matching). %These measures can be used to study the dialog system as its own ``human'' concept and how the dialog system interacts with the world, respectively. 
The next section operationalizes these classes across five metrics.

\paragraph{States and Traits} 
The state vs. trait distinction is ubiquitous in psychology, with a long history~\cite{carr1938concept}.
A standard textbook defines \textit{state measures} as thoughts, feelings, and behaviors in a specific place and time. \textit{Trait measures} are those which generalize across situations that are stable over time and systematically differ across people~\cite{zeigler2020encyclopedia}.
Emotions are states, while personalities are traits.
In relation to standard NLP tasks, past work has found stance-detection to be more trait-like while sentiment is a more state-like outcome~\cite{lynn-etal-2019-tweet}. 
It is important to distinguish the measures we use (e.g., personality), which are grounded against validated psychological instruments, from proxies for these constructs used in other works (e.g., personas). While proxy measures such as ``likes'' correlate with personality~\cite{kosinski2013private}, they are not direct assessments of the constructs.

\paragraph{Linguistic Matching}
Linguistic matching has been observed in many settings and has been shown to predict power differentials~\cite{danescu2012echoes}, relationship stability~\cite{ireland2011}, cooperation~\cite{manson2013convergence}, and empathy ratings of therapists~\cite{lord2015more}.
More generally, the psycholinguistic theory of communication accommodation has studied such unconscious matching tendencies in postures, facial expressions, pitch, pausing, length, and use of function words~\cite{giles1991contexts}. 
Besides sentence embedding similarity~\cite{zhang2021deep}, to our knowledge, such extensive matching phenomena have yet to be studied in open-domain dialog systems, despite being applied in other NLP settings~\cite{danescu2011mark,danescu-niculescu-mizil-lee-2011-chameleons}. 

\section{Psychological Metrics}

{\it Psychological metrics} operationalize the human-like measures using models trained on other data sets to predict e.g. emotion and personality. The measures include states (emotions), traits (agreeableness and empathy), and linguistic matching (emotion and style matching). 
Importantly, they were not specifically designed for evaluating dialog systems, and thus are not optimized to correlate with the gold standard human judgements in the data sets (e.g., appropriateness). 
Despite not being designed specifically for dialog evaluation, these are fundamental measures in social and psychological science, and the models employed here have been validated in previous works. 
Agreeableness and Empathy are all preexisting models trained to predict survey-based measures of their respective construct and validated in their respective studies. Similarly, language style matching is an preexisting, ``off-the-shelf'' model designed and validated in other work (see below).
The two emotion measures (emotional entropy and emotion matching) use preexisting models used to predict emotions, though the ``entropy'' and ``matching'' aspects are novel to the current work. 
Five metric scores were estimated at the turn and dialog level (depending on the metric) and then correlated with a number of crowd-sourced human judgements\footnote{\label{supplement}. See the Appendix for full details on data sets, crowd-sourced annotations, and additional experiments.}  Figure \ref{fig:enter-label} gives an example of turn- vs dialog-level evaluations.

\paragraph{Emotional Entropy} Using the NRC Hashtag Emotion Lexicon~\cite{mohammad2015using}, we estimate Plutchik's eight basic emotions: anger, anticipation, disgust, fear, joy, sadness, surprise, and trust~\cite{plutchik1980general}. 
This emotion lexicon, which is a set of weighted words for each emotion category, was automatically derived over tweets with emotion hashtags (e.g., \emph{\#anger}).
The lexicon is applied to every observation in each data set (i.e., we summed weighted word frequencies according to their weight within each emotion category) and then the entropy of the normalized emotion vector is calculated. Emotions (and, thus, emotional entropy) are state measures and can be estimated at the turn and dialog level. 

\paragraph{Agreeableness} We used a language-based personality model to estimate the agreeableness dimension of the Big Five personality traits~\cite{park2015automatic}.
This model had an out-of-sample prediction accuracy (product-moment correlation) of .35 and was built over 1-3grams and 2,000 LDA topics \cite[Latent Dirichlet Allocation;][]{blei2003latent}. Thus, for each dialog, we extracted 1-3grams and loadings for the 2,000 LDA topics and applied the pre-trained regression model,  producing an agreeableness score for each observation. We include agreeableness in our final five metrics since it outperformed the other four personality measures (openness to experience, conscientiousness, extraversion, and neuroticism) on the test data. Agreeableness (and personality, in general) is a trait measure that would typically be defined at the dialog level.

\paragraph{Empathy} We build a model to predict empathy, as measured by the Empathic Concern subscale of the Interpersonal Reactivity Index (IRI)~\cite{davis1983measuring}. We use an existing empathy data set~\cite{abdul2017recognizing,yadengiorgi2023characterizing} and build a model from 2,805 participants who shared their Facebook status data and answered the IRI questionnaire. Using 10-fold cross-validation, we predicted the empathic concern scores from a Ridge penalized linear regression using the same set of 2,000 LDA topics described above. The final model resulted in an out-of-sample product-moment correlation of 0.26. In order to obtain Empathic Concern estimates for each dialog, we extracted 2,000 LDA topic loadings for each observation and applied the pre-trained regression model. Empathic Concern is a trait-level measure. Similar to agreeableness, this would typically be defined at the dialog level. 

\paragraph{Language Style Matching} We use the definition provided by \citet{ireland2011}: 1 minus the normalized absolute difference in function word use between the agent and entity. This score was calculated for nine separate function word categories in the Linguistic Inquiry and Word Count (LIWC) dictionary~\cite{pennebaker2001linguistic}: personal pronouns, impersonal pronouns, articles, conjunctions, prepositions, auxiliary verbs, high-frequency adverbs, negations, and quantifiers. Turn- and dialog-level scores were averaged across the nine categories. This is a form of Linguistic Matching that can be measured at the turn or dialog levels.

\paragraph{Emotion Matching} Again, we use the NRC Hashtag Emotion Lexicon~\cite{mohammad2015using} and calculate the Spearman rank correlation between the agent's emotions and the prompt's emotions. Inspired by the Linguistic Style Matching metric, Emotion Matching is a form of Linguistic Matching that can be measured at the turn or dialog levels.

\begin{figure}[!tb]
\centering
\includegraphics[width=.4\textwidth]{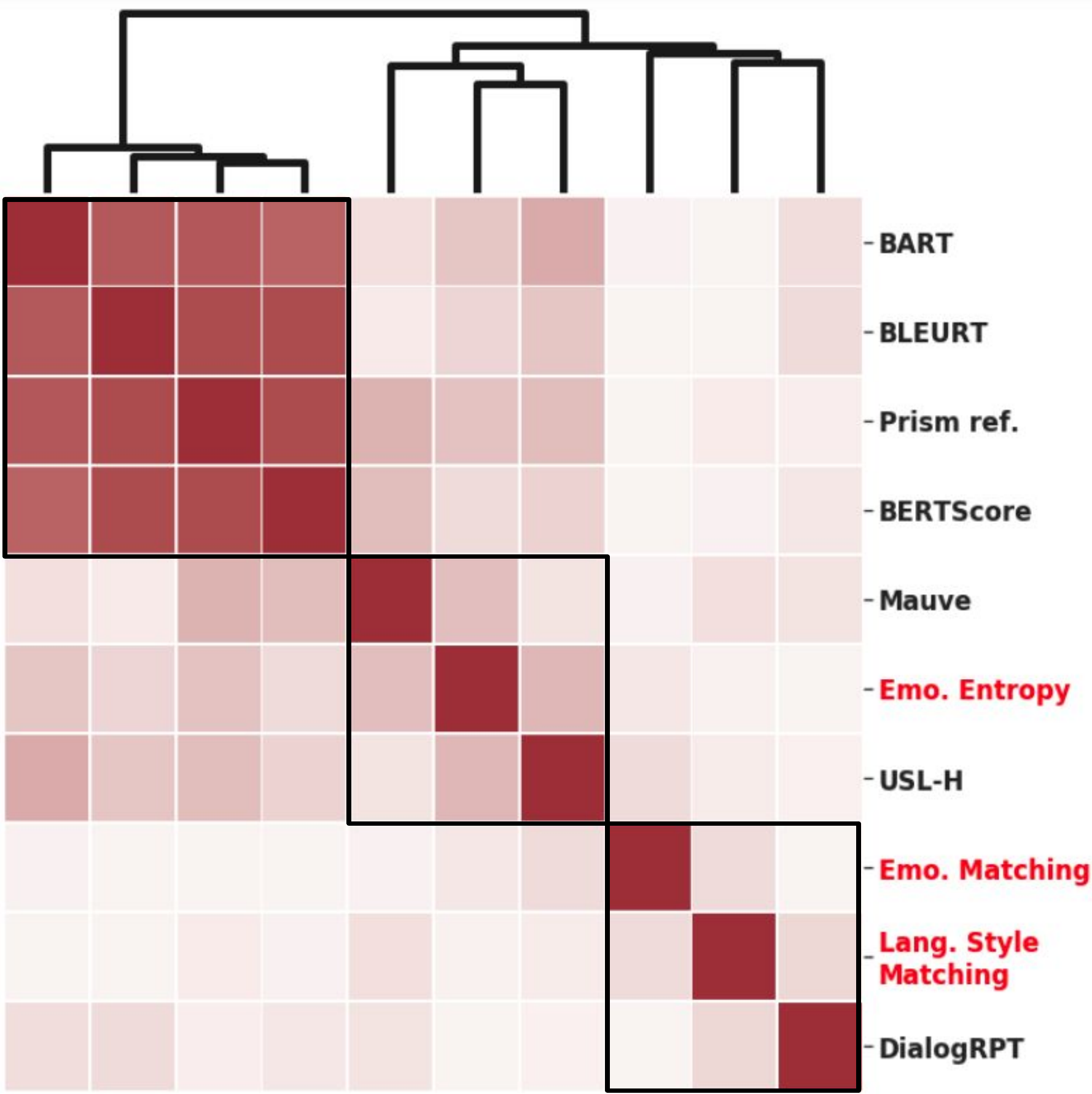}
\caption{Turn-level correlations between psychological and traditional metrics in the DSTC10 data set.  We cluster both the rows and columns based on absolute correlations. Psychological metrics are denoted in red.}
\label{fig:heatmap}
\end{figure}

\section{Data}
\label{sec:data}

To evaluate our human metrics, we collect a novel data set, the Three Bot Dialog Evaluation Corpus, from three state-of-the-art dialog systems and evaluate the dialogs at both the turn and dialog levels via crowd-sourcing (Amazon Mechanical Turk). We also evaluate our metrics on several additional open-source data sets, the DSTC10 Track 5 Test Corpus.

\subsection{Three Bot Dialog Evaluation Corpus}

Here we introduce the Three Bot Dialog Evaluation Corpus (TBD or TBD-Q1-2023; Quarter 1 of 2023). This data set consists of conversations with three chatbots: ChatGPT, GPT-3~\cite{brown2020language}, and BlenderBot~\cite{shuster2022blenderbot}. For each chatbot, we collected 21 dialogs with an average of 14.6 turns per dialog. 

The dialogs were collected from 5 different lab members (undergraduate, graduate, and faculty) having conversations with each of the three chatbots. They were collected via a Qualtrics survey instrument. Participants were instructed to conduct a 15 to 30 turn conversation with model.

We then collect human judgments at both the turn and dialog level for each conversation in the data set, using Amazon Mechanical Turk (AMT). Our annotators are restricted to the users with location US, >97\% approval rate, >1000 HITs done, and a convenience pool of workers used for NLP evaluation tasks. We included text-based attention checks at the dialogue-level as well as an annotator agreement (both with an expert as well as between crowd workers) time-based filters on the turn-level. We used 5 annotators for the dialog-level and 3 annotators for the turn-level annotations. Annotators were paid \$15/hour.

At the turn level, we ask crowd workers to evaluate across several dimensions: appropriateness, content, grammar, and relevance. At the dialog level, we ask crowd workers to evaluate the conversation for coherence, informativeness, likability, and overall (exact evaluation question text is included in the Appendix).  The linear Krippendorff’s alpha (averaged across all dimensions) for dialog-level was 0.45 and turn-level was 0.63. 

Given the lack of references included in this data set, we can only apply the reference-free traditional metrics: DialogRPT, Mauve, and USL-H. Further, since there are evaluations at both the turn- and dialog-level, we can evaluate all psychological metrics: agreeableness, empathy, emotional entropy, emotion matching, and language style matching.

\subsection{DSTC10 Track 5 Turn-level Test Corpus} In order to further evaluate our human metrics, we use a test corpus from The Tenth Dialog System Technology Challenge (DSTC10) Track 5 Automatic Evaluation and Moderation of Open-domain Dialogue Systems~\cite{zhang2021auteval}. This evaluation data set combined five \emph{turn level} data sets into a single data set: Topical-DTSC10, Persona-DSTC10, CHANEL-JSALT-2020 (JSALT; ~\citealt{kong2019subjective}), Neural Conversation Model (NCM; ~\citealt{vinyals2015neural}), English As a Second Language (ESL; ~\citealt{lee2020evaluation}).\footref{supplement} Since this data set is available at the turn level, we evaluate our three turn-level metrics (emotional entropy, emotion matching, and language style matching) as well as the traditional metrics. 

\paragraph{DSTC10 Track 5 Dialog-level Corpus} 
We use the dialog-level data set from the DSTC10 shared task: FED-Dial~\cite{mehri2020unsupervised}. As this corpus is evaluated (via human judgements) at the dialog-level, we apply our two dialog-level psychological metrics: agreeableness and empathy. Traditional metrics are averaged from the turn to the dialog. 
% We use the two dialog-level data sets from the DSTC10 shared task: Persona-See~\cite{see2019makes} FED-Dial~\cite{mehri2020unsupervised}. Here, these are combined into a single evaluation corpus. 

\begin{table*}[ht]
\centering
\resizebox{.98\textwidth}{!}{ 
\begin{tabular}{clccccccccc}\toprule
& \multicolumn{1}{c}{\multirow{2}{*}{}} & \multirow{2}{*}{\begin{tabular}[c]{@{}c@{}}Traditional \\ Metric Alone\end{tabular}} &  \multicolumn{2}{c}{Emo. Entropy} & \multicolumn{2}{c}{Emo. Matching} & \multicolumn{2}{c}{Lang. Style Matching} & \multicolumn{2}{c}{All Psych.} \\ \cmidrule(lr){4-5} \cmidrule(lr){6-7} \cmidrule(lr){8-9} \cmidrule(lr){10-11} 
& \multicolumn{1}{c}{} &  & P & P+T & P & P+T & P & P+T & P & P+T \\ \midrule
\multirow{3}*{\rotatebox{90}{TBD}} & DialogRPT & .133 & .017 & .138 & .014 & .138 & .001 & .135 & .031 & .144$^{**}$ \\
 & Mauve & -.001 & .017 & .016$^{*}$ & .014 & .013$^{*}$ & .001 & .000 & .031 & .031$^{**}$ \\
 & USL-H & .000 & .017 & .017$^{*}$ & .014 & .014$^{**}$ & .001 & .001 & .031 & .031$^{***}$\\
\midrule
\multirow{7}*{\rotatebox{90}{DSTC10}}\hspace{2mm}\multirow{7}*{\rotatebox{90}{Turn-level}} & BARTScore &   .072 & .097 & .138$^{***}$ & .001 & .073 & .056 & .128$^{***}$ & .148 & .190$^{***}$ \\
 & BERTScore & .048 & .097 & .128$^{***}$ & .001 & .049$^{*}$ & .056 & .101$^{***}$ & .148 & .178$^{***}$ \\
 & BLEURT & .031 & .097 & .113$^{***}$ & .001 & .032 & .056 & .086$^{***}$ & .148 & .165$^{***}$ \\
 & DialogRPT & .188 & .097  & .289$^{***}$ & .001 & .190$^{**}$ & .056 & .218$^{***}$ & .148 & .315$^{***}$ \\
 & Mauve & .095 & .097 & .152$^{***}$ & .001 & .096 & .056 & .135$^{***}$ & .148 & .192$^{***}$ \\
 & Prism ref. & .102 & .097 & .159$^{***}$ & .001 & .104 & .056 & .148$^{***}$ & .132 & .205$^{***}$ \\
 & USL-H & .104 &  .097 & .155$^{***}$ & .001 & .104 & .056 & .170$^{***}$ & .148 & .215$^{***}$ \\

\bottomrule
\end{tabular}
}
\caption{Turn-level Results, predicting the ``Appropriateness'' human judgement: Reported linear regression adjusted $R^2$ where $P$ contains the psychological metrics as the independent variable and $P+T$ contains both the psychological and traditional metrics as independent variables. Bonferroni corrected significance level: $^{***}$ p < 0.001, $^{**}$  p < 0.01, $^{*}$ p < 0.05}
\label{tab:turn-level results}
\end{table*}

\section{Traditional metrics}

We compare the psychological metrics to seven metrics traditionally used to evaluate dialog systems. All metrics are turn-level metrics and, when used at the dialog-level, are averages across all turns within a given dialog. 

\textbf{BARTScore} is a metric that evaluates generated text using a pre-trained BART encoder-decoder model~\cite{yuan2021bartscore}. It formulates the generated text evaluation as a text generation problem by directly evaluating text with the probability of being generated from or generating other textual inputs and outputs.

\textbf{BERTScore} is an evaluation metric for text generation that computes the similarity of two sentences as a sum of the cosine similarities between pre-trained BERT contextual embeddings~\cite{zhang2019bertscore}. For dialog systems, it computes the F1 scores by matching token embeddings in the human reference and system response.

\textbf{BLEURT} is a text generation evaluation metric based on BERT that can model human judgements~\cite{sellam2020bleurt}. This uses a pre-training scheme on BERT with synthetic data and fine-tunes it to predict a human score with a mean squared error (MSE) loss when applied to dialog systems.

\textbf{DialogRPT} is an ensemble model consisting of GPT-2 based models trained on human feedback data for tasks predicting the feedback and how human-like responses are~\cite{gao2020dialogue}. 

\textbf{Mauve} measures differences in neural and human written text via  Kullback-Leibler divergence~\cite{pillutla2021mauve}.

\textbf{Prism} is a machine translation evaluation framework that uses a sequence-to-sequence paraphraser to score outputs conditioned on a human reference~\cite{thompson-post-2020-automatic}. This uses a multilingual neural machine translation (NMT) model as a zero-shot paraphraser which was trained by treating the paraphrasing as a translation task.

\textbf{USL-H} is a dialog evaluation metric that uses a composition of measurements for understandability, sensibleness, and likeability~\cite{phy2020deconstruct}. This uses models trained for valid utterance prediction (VUP) to determine validity, along with next sentence prediction (NSP) and masked language modeling (MLM) models to measure sensibleness and likelihood of a response.

% \paragraph{Traditional metrics} We use 6 common dialog system metrics: BARTScore~\cite{yuan2021bartscore}, BERTScore~\cite{zhang2019bertscore}, BLEURT~\cite{sellam2020bleurt}, DialogRPT~\cite{gao2020dialogue}, Prism~\cite{thompson-post-2020-automatic}, Mauve~\cite{pillutla2021mauve}, and USL-H~\cite{phy2020deconstruct}.\footref{supplement} 
% TBD-Q1-2023 is evaluated using DialogRPT, Mauve, and USL-H only.
% All metrics, with the exception of Mauve, are used to evaluate the DSTC10 Track 5 Test Corpus.
% As these are all turn-level metrics, we average metrics across turns to create dialog-level scores when evaluating TBD-Q1-2023 at the dialog-level.%\joao{Do we need to say anything more here or above with the data sets?} NO

\begin{table*}[!ht]
\centering
\begin{tabular}{clccccccc}\toprule
& \multicolumn{1}{c}{\multirow{2}{*}{}} & \multirow{2}{*}{\begin{tabular}[c]{@{}c@{}}Traditional \\ Metric Alone\end{tabular}} &  \multicolumn{2}{c}{Agreeableness} & \multicolumn{2}{c}{Empathy} & \multicolumn{2}{c}{All Psych.} \\ \cmidrule(lr){4-5} \cmidrule(lr){6-7} \cmidrule(lr){8-9} 
& \multicolumn{1}{c}{} &  & P & P+T & P & P+T & P & P+T \\ \midrule
\multirow{3}*{\rotatebox{90}{TBD}} & DialogRPT & .180 & .094 & .175 & .031 & .168 & .089 & .161 \\
 & Mauve & .091 & .094 & .120 & .031 & .084 & .089 & .106 \\
 & USL-H & .002 & .094 & .120 & .031 & .089 & .089 & .146\\ \midrule
\multirow{3}*{\rotatebox{90}{DSTC10}}\hspace{2mm}\multirow{3}*{\rotatebox{90}{Dialog}} & DialogRPT & .010 & .044 & .049 & .000 & .011 & .040 & .046 \\
 & Mauve & .012 & .044 & .056 & .000 & .013 & .040 & .054 \\
 & USL-H & .108 & .044 & .140 & .000 & .105 & .040 & .135\\
\bottomrule
\end{tabular}
\caption{Dialog-level results, predicting the ``Overall'' human judgement: Reported linear regression adjusted $R^2$ where $P$ contains the psychological metrics as the independent variable and $P+T$ contains both the psychological and traditional metrics as independent variables. Bonferroni corrected significance level: $^{***}$ p < 0.001, $^{**}$  p < 0.01, $^{*}$ p < 0.05
}
\label{tab:dialog-level results}
\end{table*}

\section{Evaluation}
\label{sec:evaluation}

To evaluate the psychological metrics, we proceed in three steps: (1) we correlate both the psychological and traditional metrics in order to identify potential similarities between the metrics; (2) we use both the psychological and traditional metrics to predict human judgements (both at the turn- and dialog-levels); and (3) we use the psychological and traditional metrics to characterize the dialog systems in the absence of human judgements, in order to gain insights into each systems' conversational capabilities.

\subsection{Metric Correlations} 
First, we compute pairwise correlations (product-moment correlations) between the psychological and traditional metrics. These correlations are visualized via a heat map, where both rows and columns are clustered via their absolute effect size. This clustering allows us to identify correlational patterns between the metrics, helping to identify how the psychological metrics are related to the traditional metrics.  

\subsection{Correlation with Human Judgements} 
We create three models which contain varying sets of independent variables: (1) the traditional metric (``T''), (2) the psychological metric (``P''), and (3) both the psychological and traditional metrics together (``P+T''). In all models, the dependent variable is the median of crowd-sourced annotation.\footref{supplement} The human judgement is \emph{Appropriateness} (``The response is appropriate given the preceding dialog.'') for the turn-level evaluations and \emph{Overall} (``Overall impression of the dialog'') for the dialog-level evaluations. Additionally, all variables are mean-centered and standardized so that the resulting standard deviation is equal to 1. We report model fit via adjusted $R^2$. We also perform a paired t-test between the mean absolute residuals of the ``T'' and ``P+T'' models to see if the psychological metrics add significant predictive value above the traditional metrics alone. We then apply a Bonferroni correction to compensate for the large number of comparisons~\cite{armstrong2014use}.

\subsection{Characterizing Dialog Systems without Human Judgements} 
Here we characterize the three systems in the Three Bot Dialog Evaluation Corpus (ChatGPT, BlenderBot, and GPT-3) using the psychological and traditional methods alone (i.e., no human judgement). This is done to highlight the differences between the three systems. Turn-level metrics are averaged across dialogs and further averaged across dialogs within a given system. Dialog-level metrics are simply averaged across dialogs within a system. Here we highlight the fact that the psychological metrics are interpretable and can serve as stand-alone evaluations of dialog systems. Given that these metrics are automatically applied and can thus be applied at scale across large data sets, they may open up the possibility of cheaper and less time-consuming evaluations that can be used alongside human judgements. To aid visualization, scores are normalized to be between 0 and 1.

\begin{figure*}[!ht]
     \centering
     \begin{subfigure}[]{.99\textwidth}
         \centering
         \includegraphics[width=\textwidth]{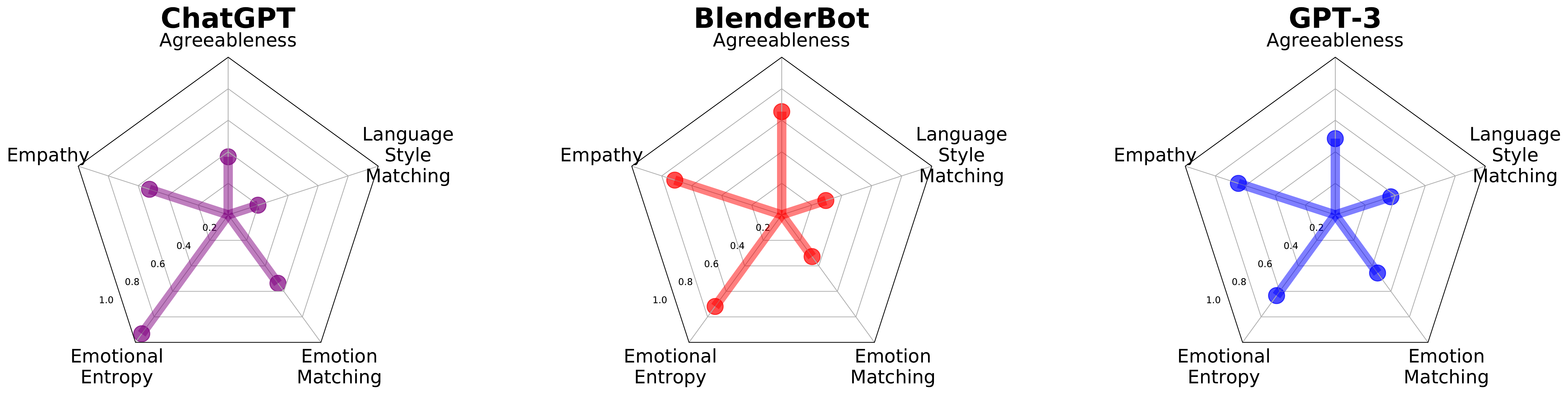}
         \caption{Psychological metrics}
         \label{fig:psych TBD}
     \end{subfigure}
     \\ \vspace{3mm}
     \begin{subfigure}[]{0.99\textwidth}
         \centering
         \includegraphics[width=\textwidth]{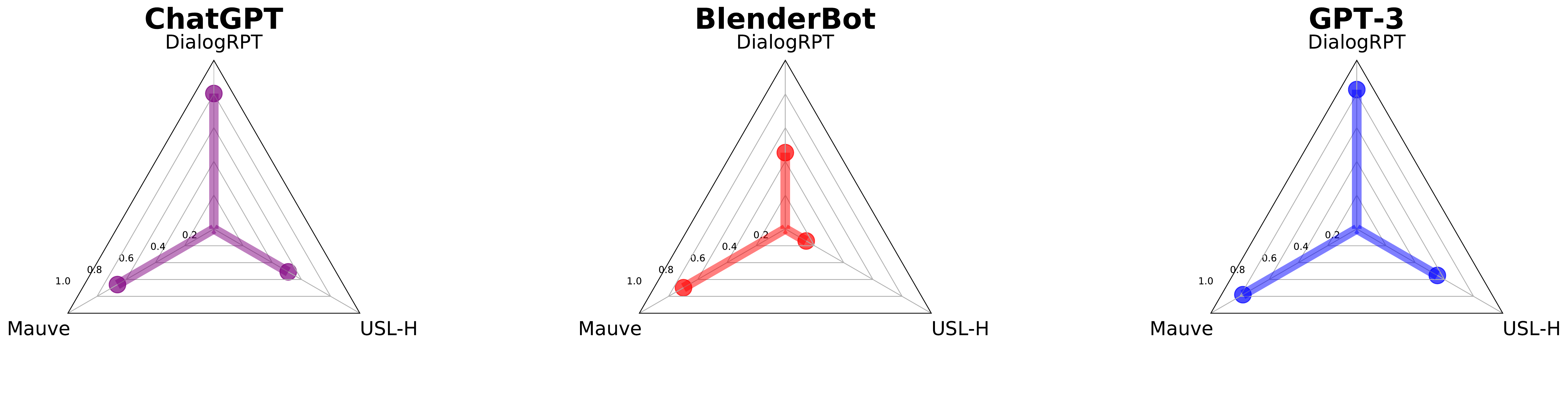}
         \caption{Traditional metrics}
         \label{fig:trad TDB}
     \end{subfigure}
     \caption{Comparison of (a) psychological  and (b) traditional metrics across the three dialog systems in the Three Bot Dialog Evaluation Corpus: ChatGPT (purple), BlenderBot (red), and GPT-3 (blue). Scores are normalized to be between 0 and 1 in order to aid visualization.}
     \label{fig:metrics alone}
\end{figure*}

\section{Results}
\label{sec:results}

Figure \ref{fig:heatmap} shows the clustered correlations between the \textit{psychological} and \textit{traditional} metrics on the DSTC10 Track 5 Test Corpus. Three distinct clusters appear: (1) BARTScore, BERTScore, BLEURT, and Prism ref.; (2) Mauve, Emotional Entropy, and  USL-H; and (3) Emotion Matching, Language Style Matching, and DialogRPT. As expected, all reference-based contextualized embedding methods cluster together. The effect sizes in the second and third clusters are smaller than the first cluster, suggesting these are less coherent clusters. Indeed, as seen in the dendrogram, the second and third clusters and be combined and are distinct from the first cluster. %\joao{Can you write a sentence on Cluster (1)? Would you expect these to cluster?} 

Table \ref{tab:turn-level results} shows the comparison between the psychological and traditional metrics when predicting the turn-level human judgements. Several state-of-the-art traditional metrics performed well, such as BARTScore and BLEURT. While the psychological metrics did not perform as well, we see that emotional entropy, emotion matching, and language style matching all increase predictive accuracy when combined with the traditional metrics. Table \ref{tab:dialog-level results} shows the results of the dialog-level analysis, predicting the Overall annotation. (See Appendix Tables \ref{tab:dialog-level results coherence}, \ref{tab:dialog-level results informativeness}, and \ref{tab:dialog-level results likability} for coherence, informativeness, and likability results.) Due to the small sample size of TBD (63 dialogs) and DTSC10 Dialog-level Corpus (125 dialogs), we do not have statistical power to identify differences between the psychological and automatic metrics. That said, agreeableness performed at the same level or above two of the traditional metrics in each data set.

Taken together, the psychological metrics were not highly predictive alone when compared to state-of-the-art metrics (which is expected since the psychological metrics are not specific for dialog evaluations), yet they are capturing unique, relevant signal for dialog quality. Similar results hold across an additional 10 out of 12 open-domain dialog evaluation data sets in the Appendix.\footref{supplement} 

In Figure \ref{fig:metrics alone}(a), we see ChatGPT, BlenderBot, and GPT-3 evaluated across the psychological metrics. 
We see that ChatGPT lacks both empathy and personality, BlenderBot is high on empathy, agreeableness, and highest emotional variation, and GPT-3 is high on empathy and low on agreeableness. All systems are low on language-style matching. 
The traditional metrics in Figure \ref{fig:metrics alone}(b) show (1) little variation in Mauve, (2) ChaptGPT and GPT-3 are similarly high in DialogRPT, and (3) BlenderBot and GPT-3 are both higher on Mauve than ChatGPT. Despite these differences, all three systems look relatively similar across these dimensions.

% \begin{figure*}[!ht]
%     \includegraphics[width=0.99\textwidth]{figures/comparison-graphs.pdf}
%      \caption{Comparison of traditional and psychological metrics across the three dialog systems in the Three Bot Dialog Evaluation Corpus: ChatGPT, BlenderBot, and GPT-3. Scores are normalized to be between 0 and 1 in order to aid visualization.}
%      \label{fig:metrics alone}
% \end{figure*}

\section{Conclusions}
\label{sec:conclusions}

This paper proposes several psychologically-grounded measures for evaluating open-domain dialog systems. Our metrics evaluate dialog systems from a human lens, considering both trait and state trade-offs (standard measures of human constructs) and linguistic matching (indicators of social relationships and interactions). 

We evaluate all five of our proposed metrics, examining trait-level features (agreeableness and empathy), state-level variation (emotional entropy), and linguistic matching (style and emotion matching). We also compare against state-of-the-art traditional metrics across multiple data sets and show that the psychological metrics (1) do not correlate with traditional metrics and (2) lead to increased accuracy when predicting gold standard human judgements. These results indicate our psychological metrics are picking up on unique signal when evaluating open-domain dialog systems. 

Finally, we characterize dialog systems using both traditional and psychological metrics. Our findings suggest that ChatGPT, in particular, lacks empathy and personality, which are fundamental human traits, despite claims ChatGPT has human-like qualities, such as a theory of mind~\cite{kosinski2023theory}. Therefore a more careful evaluation of its role in sensitive use cases like therapy~\cite{stade2023artificial} is needed. 
The traditional metrics measure concepts such as plausibility (Mauve) or understandability (USL). Thus, it is hard to interpret these results in their relationships to human communication. For example, agreeableness is associated with cooperation and trustworthiness~\cite{stavrova2022effects}, but it is unclear how plausibility or understandability are related to similar concepts. 

Current large language models such as GPT-4 perform fantastically well when evaluated at the utterance level. However, they are much weaker at the conversation and person level. Researchers and companies are now trying to build chatbots that have consistent personalities (``personas'') and can carry out conversations with internal structure such as introductory small talk (``How is your week going''), or concluding formalities (``It was wonderful working with you.''). Evaluating chatbots from this perspective requires better metrics; traditional metrics are often too weak to distinguish between modern dialog agents, as all current LLMs exhibit human-level fluency and strong topic knowledge. In contrast, our psychologically-grounded metrics show large and informative differences between agents, helping to better characterize their performance.

\section*{Ethical Considerations} There are several ethical considerations when constructing and evaluating dialog systems, many of which have been outlined by Roller et al.~\cite{roller2021recipes}. These include privacy (since online dialog may contain sensitive information), toxic and offensive content, and, on the part of the researcher, openness to sharing findings.  
With regard to the current work, imparting systems with human qualities such as personality and socio-demographics must be handled with the utmost sensitivity. Biases in training data, misclassifications in downstream tasks, and reliance on outdated social constructs (i.e., binary gender) are just a few examples of how automated systems can fail and further marginalize vulnerable populations~\cite{shah2020predictive,xudetoxifying,gonen2019lipstick}. 
Specifically, the models used in this study (e.g., empathy and agreeableness) are trained on majority U.S. and monolingual English-speaking populations and may fail to generalize to minority or non-US populations.
On the other hand, the alternative also suffers from similar concerns, namely that dialog systems may exhibit extremely limited variation in such traits. One could imagine a similar situation to the so-called ``Wall Street Journal effect'' (i.e., part-of-speech taggers are only accurate when applied to language written by white men; ~\cite{hovy2015tagging}), where dialog system only converse like middle-aged white men.

It is also important to note that while the proposed classes of metrics (i.e., states/traits and linguistic matching) may be desirable in the context of ``human-like'' measures, the examples used in the paper (e.g., agreeableness) may not. When presented with a toxic prompt, an agreeable or style-matching dialog system will only reinforce the toxicity by agreeing with or matching the prompt, while embedding systems with social norms may help alleviate such issues~\cite{kim2022prosocialdialog}. In general, more human-like dialog systems, as enabled by this approach, can be used both for good (better support for mental health) and for evil (more effective deception and misinformation). Thus, care must be taken when choosing constructs to be embedded in dialog systems.  

Finally, it is important not to anthropomorphize dialog systems as this can lead to transparency and trust issues, especially in high-stakes settings (see ~\citet{abercrombie2023mirages} for an in-depth discussion). While we are suggesting metrics grounded in human psychology for evaluation, we do not mean to imply that these systems are human, human-like, or should be thought of as human. 

\section*{Limitations}
While we have attempted to evaluate our metrics on a large number of public data sets and compare them against many state-of-the-art metrics, there are a number of limitations. First, the psychological metrics are not developed for dialog system evaluations and may fail to capture the nuances of this domain. For example, the agreeableness model was trained on lifetime post histories from Facebook users, and thus one may not expect this to work well on short responses within a dialog or even conversations in general. Next, the specific metrics proposed in this paper (e.g., agreeableness and empathy) are just five examples of psychologically grounded measures that could be applied in this setting. We do not claim to have attempted a thorough investigation across all possible (or even a large number of) psychological metrics. Finally, there is no reason to expect the proposed psychological metrics to correlate with human judgments. For example, it is not immediately clear that emotional entropy should correlate with either ``appropriateness'' or ``relevance''. %Given psychologically related psychological annotations, the psychological metrics may or may not correlate with or add value to the traditional metrics. 

%Bibliography

\bibliography{references}  
\bibliographystyle{acl_natbib}


\appendix

\section{TBD Human Judgement Evaluations}
The TBD-Q1-2023 was evaluated at both the dialogue- and turn-level by crowd workers on Amazon Mechanical Turk. Each dialogue was evaluted across 4 dimenions: coherence, informativeness, likability, and overall. 
Coherence (or Understanding) is a 5 item Likert scale with 1 representing ``The entire conversation is incomprehensible'' and 5 representing ``The dialogue is very coherent and all the information conveyed is consistent''. 
Informativeness is a 5 item Likert scale with 1 representing ``There is barely any information content in the dialogue, such as generic utterances, perfunctory responses, and repetition. Often the utterances in the dialogue are short. Dialogues that receive a rating of 1 for understanding/coherence'' and 5 representing ``Most of the utterances in the dialogue are long sentences with high information content, and all the information is correct''. 
Likability (or Engagingness) is a 5 item Likert scale with 1 representing ``The content of the conversation is unattractive, and I don't know how to continue the conversation; dialogues receive a rating of 1 for understanding/coherence'' and 5 representing ``The conversation is extremely attractive and I am eager to continue it''. 
Overall is a 5 item Likert scale with 1 representing ``The overall quality is very low, the conversation is not fluent and there is no information'' and 5 representing ``The overall quality is excellent, the conversation is very smooth, the amount of information content is very high with great engagingness, it's a very good response''. 

The turn-level was evaluated for Grammatical Correctness (``The quality of the English grammar''), Appropriateness (``The response is appropriate given the preceding turn (Note: The appropriateness of a response is very subjective''), Content richness (``The response is informative, containing long sentences that include various entities (such as names of people, names of places or times), conceptual words (sky, dust, sorrow, etc.) or descriptive/emotional words (It hurts me, Lovely, etc.)''), and Relevance (``The response is related to the context of the dialogue and is good and smooth'').
All items were on a 1 to 5 Likert scale, with 1 being lowest and 5 being highest (e.g., 1 = no grammatical correctness). 

\section{Additional TBD Dialogue-level Evaluations}

Tables \ref{tab:dialog-level results coherence}, \ref{tab:dialog-level results informativeness}, and \ref{tab:dialog-level results likability} show the results of our human metrics predicting the Coherence, Informativeness, and Likability crowd sourced dialogue-level annotations on the TBD-Q1-2023 data set. 

\begin{table*}[ht]
\centering
\begin{tabular}{clccccccc}\toprule
& \multicolumn{1}{c}{\multirow{2}{*}{}} & \multirow{2}{*}{\begin{tabular}[c]{@{}c@{}}Traditional \\ Metric Alone\end{tabular}} &  \multicolumn{2}{c}{Agreeableness} & \multicolumn{2}{c}{Empathy} & \multicolumn{2}{c}{All Psych.} \\ \cmidrule(lr){4-5} \cmidrule(lr){6-7} \cmidrule(lr){8-9} 
& \multicolumn{1}{c}{} &  & P & P+T & P & P+T & P & P+T \\ \midrule
\multirow{3}*{\rotatebox{90}{TBD}} & DialogRPT & .186 & .115 & .188 & -.012 & .182 & .102 & .190 \\
 & Mauve & .081 & .115 & .128 & -.012 & .068 & .102 & .124 \\
 & USL-H & .048 & .115 & .183 & -.012 & .073 & .102 & .173\\
\bottomrule
\end{tabular}
\caption{Dialogue-level results predicting the Coherence rating: Reported linear regression adjusted $R^2$ where $P$ contains the psychological metrics as the independent variable and $P+T$ contains both the psychological and traditional metrics as independent variables. %Benjamini-Hochberg corrected significance level: $^{***}$ p < 0.001, $^{**}$  p < 0.01, $^{*}$ p < 0.05
}
\label{tab:dialog-level results coherence}
\end{table*}

\begin{table*}[ht]
\centering
\begin{tabular}{clccccccc}\toprule
& \multicolumn{1}{c}{\multirow{2}{*}{}} & \multirow{2}{*}{\begin{tabular}[c]{@{}c@{}}Traditional \\ Metric Alone\end{tabular}} &  \multicolumn{2}{c}{Agreeableness} & \multicolumn{2}{c}{Empathy} & \multicolumn{2}{c}{All Psych.} \\ \cmidrule(lr){4-5} \cmidrule(lr){6-7} \cmidrule(lr){8-9} 
& \multicolumn{1}{c}{} &  & P & P+T & P & P+T & P & P+T \\ \midrule
\multirow{3}*{\rotatebox{90}{TBD}} & DialogRPT & .230 & .112 & .224 & .061 & .225 & .121 & .216 \\
 & Mauve & .141 & .112 & .166 & .061 & .142 & .121 & .159 \\
 & USL-H & -.005 & .112 & .116 & .061 & .110 & .121 & .161\\
\bottomrule
\end{tabular}
\caption{Dialogue-level results predicting the Informativeness rating: Reported linear regression adjusted $R^2$ where $P$ contains the psychological metrics as the independent variable and $P+T$ contains both the psychological and traditional metrics as independent variables. %Benjamini-Hochberg corrected significance level: $^{***}$ p < 0.001, $^{**}$  p < 0.01, $^{*}$ p < 0.05
}
\label{tab:dialog-level results informativeness}
\end{table*}

\begin{table*}[ht]
\centering
\begin{tabular}{clccccccc}\toprule
& \multicolumn{1}{c}{\multirow{2}{*}{}} & \multirow{2}{*}{\begin{tabular}[c]{@{}c@{}}Traditional \\ Metric Alone\end{tabular}} &  \multicolumn{2}{c}{Agreeableness} & \multicolumn{2}{c}{Empathy} & \multicolumn{2}{c}{All Psych.} \\ \cmidrule(lr){4-5} \cmidrule(lr){6-7} \cmidrule(lr){8-9} 
& \multicolumn{1}{c}{} &  & P & P+T & P & P+T & P & P+T \\ \midrule
\multirow{3}*{\rotatebox{90}{TBD}} & DialogRPT & .081 & .043 & .071 & .054 & .090 & .063 & .076 \\
 & Mauve & .021 & .043 & .035 & .054 & .046 & .063 & .047 \\
 & USL-H & -.014 & .043 & .033 & .054 & .077 & .063 & .081\\
\bottomrule
\end{tabular}
\caption{Dialogue-level results predicting the Likability rating: Reported linear regression adjusted $R^2$ where $P$ contains the psychological metrics as the independent variable and $P+T$ contains both the psychological and traditional metrics as independent variables. %Benjamini-Hochberg corrected significance level: $^{***}$ p < 0.001, $^{**}$  p < 0.01, $^{*}$ p < 0.05
}
\label{tab:dialog-level results likability}
\end{table*}

%%%%%%%%%%%%%%%%
%%%%%%%%%%%%%%%% DATA SETS
%%%%%%%%%%%%%%%%
\section{Additional Data Sets}
% needed JSALT
% done
\textbf{DSTC6 (D6)} is dialogue data collected from Twitter users for customer service for 40,000 context-response pairs~\cite{hori2017end}. The dialogue context was evaluated using 10 Turkers on a 5 point Likert scale based on the relevance of the response. 

% done
\textbf{DSTC7 (D7)} is conversation data extracted from Reddit conversation threads~\cite{galley2019grounded}. The dataset contained 3 million conversational responses and 20 million facts. The dialogue context was evaluated by crowdsourced annotators using a 5 point Likert scale based on the relevance and interest of the response.

% % missing
% \textbf{ConvAI2-Eval (EC)} is a persona-conditional dialogue dataset based on the dialogues from the test set of the ConvAI2 dataset through GRADE metric annotation~\cite{huang2020grade}. The context-response pairs were evaluated by Turkers using a 1 to 5 scale based on coherence.

% % missing
% \textbf{DailyDialog-Eval (ED)} is a dialogue dataset based on the dialogues from the test set of the DailyDialog dataset through GRADE metric annotation~\cite{huang2020grade}. The context-response pairs were evaluated by Turkers using a 1 to 5 scale based on coherence.

% % missing
% \textbf{Empathetic-Eval (EE)} is a dialogue dataset based on the dialogues from the test set of the EmpatheticDialogues dataset through GRADE metric annotation~\cite{huang2020grade}.The context-response pairs were evaluated by Turkers using a 1 to 5 scale based on coherence.

% done
\textbf{English As a Second Language (ESL)} consists of 200 different three turn dialogue segments from an English learning site~\cite{zhang2021auteval}. This dataset consists of 21 comparisons across 5 dialogue systems with a human baseline over 13K judgements.

% % missing
% \textbf{FED-Conversation (FC)} is a conversation dataset collected from the Meena and Mitsuku open-domain dialogue systems~\cite{mehri2020unsupervised}. Annotations were performed using Turkers at the dialogue level on 11 criteria.

% % missing
% \textbf{FED-Turn (FT)} uses the same conversation data as the FED-Conversation dataset~\cite{mehri2020unsupervised}. Annotations were performed using Turkers at the turn level on 9 criteria.

% done
\textbf{DailyDialog (GD)} is a dialogue dataset constructed using 100 dialogue contexts from the test set of the DailyDialog dataset~\cite{gupta2019investigating}. The context-response pairs were annotated by Tukers using a 1 to 5 scale based on appropriateness.

% done
\textbf{HUMOD (HU)} is a multi-turn movie dialogue dataset created from the Cornell Movie-Dialogs Corpus~\cite{merdivan2020human}. This dataset is human annotated on a 1 to 5 scale based on the relevance of human generated responses to the context of a fictional conversation on the movie script.

% done
\textbf{Neural Conversation Model (NCM)} consists of 200 hand-crafted single turn prompts originally from the IT Helpdesk Troubleshooting dataset~\cite{zhang2021auteval}. This dataset consists of 59 comparisons across 11 dialogue systems with over 33K pairwise comparisons.

% % missing
% \textbf{Persona-Chatlog (PC)} consists of data obtained from 3,316 conversations across 26 models and a human agent~\cite{see2019makes}. Annotations were performed at the dialogue level for each model on a 4 point Likert scale based on avoiding repetition, interestingness, making sense, fluency, listening, inquisitiveness, humanness, and engagingness.

% done
\textbf{Persona-DSTC10 (PD10)} is an evaluation dataset for the DSTC10 challenge constructed from a sample of 500 dialogue segments from the PersonaChat dataset~\cite{zhang2021auteval}. A total of 4,500 context-response pairs were rated using an automatic dialogue response evaluator.

% done
\textbf{Topical-DTSC10 (TD10)} is an evaluation dataset for the DSTC10 challenge constructed from a sample of 500 dialogue segments from the TopicalChat dataset~\cite{zhang2021auteval}. A total of 5,000 context-response pairs were evaluated using an automatic dialogue response evaluator.

% done
\textbf{TopicalChat-USR (TP)} is a human evaluation dataset developed from the Topical-Chat dataset through the USR metric annotation~\cite{mehri2020usr}. The context-response pairs were annotated by Turkers using a different scales based on qualities of understanding (0-1), natural (1-3), maintains context (1-3), interesting (1-3), uses knowledge (0-1), and overall quality (1-5).

% done
\textbf{PersonaChat-USR (UP)} is a human evaluation dataset developed from the PersonaChat dataset the same way as TopicalChat-USR~\cite{mehri2020usr}. The context-response pairs are annotated with the same USR annotation scheme as TopicalChat-USR using the same qualities and scales.

% done
\textbf{DailyDialog (ZD)} is a dialogue dataset constructed using 100 dialogue contexts from the test set of the DailyDialog dataset~\cite{zhao2020designing}. The context-response pairs were annotated by Turkers using a 5 point Likert scale based on appropriateness, language usage, relevance, and context.

% done
\textbf{PersonaChat (ZP)} is a dialogue dataset conisisting of context-response pairs collected from the test set of the PersonaChat dataset~\cite{zhao2020designing}. The appropriateness quality of the response were annotated by Turkers for each context.

%%%%%%%%%%%%%%%%
%%%%%%%%%%%%%%%% HUMAN EVALUATIONS
%%%%%%%%%%%%%%%%
\section{Human Judgements}

Table \ref{tab:human evaluations} lists the human judgements used across the additional data sets used in the supplement. Each turn or dialog may have been annotated by multiple crowd-workers, depending on the data set (e.g., a single prompt may have multiple crowd-sourced evaluations for Appropriateness). The median evaluation is then used as the gold standard for each unit in the data set.

% overall for dialog-level, and appropriateness for the turn-level

\begin{table*}[ht]
\centering
\resizebox{.98\textwidth}{!}{ 
\begin{tabular}{lccc}\toprule
Judgement & Question Text & Likert Scale & Data Sets \\ \hline
 Appropriateness & The response is appropriate given the preceding dialogue.  & 1-5 & ESL, NCM, PD10, TD10, ZD, ZP \\ 
 Relevance       & The response content is related to the preceding dialogue. & 1-5 & EC, ED, EE, HU \\ 
 Enjoy           & How much did you enjoy talking to this user? & 1-4 & PC \\ 
 Overall         & What is your overall impression of the quality of this utterance? & 1-5 & D6, D7, GD, FC, FT, TP, UP  \\ 
\bottomrule
\end{tabular}
}
\caption{Human judgements for the supplemental turn-level data sets.}
\label{tab:human evaluations}
\end{table*}

\section{Results}

Tables \ref{tab:full results DSTC6} through \ref{tab:full results personachat zp} contain results for each data set. All tables report adjusted $R^2$ from a linear regression model whose dependent variable is the human evaluation metric (described above). We create three models which contain varying sets of independent variables: (1) the traditional metric alone (``Traditional Metric Alone''), (2) the psychological metric alone (``P''), and (3) both the psychological and traditional metrics together (``P+T''). In all models, the independent variables are mean centered and standardized, so that the resulting standard deviation is equal to 1. Note that ``All Psych.'' contains all five psychological metrics: agreeableness, empathy, emotional entropy, emotion matching, and language style matching. 

% \begin{table*}[ht]
% \centering
% \resizebox{.98\textwidth}{!}{ 
% \begin{tabular}{lccccccccccccc}\toprule
%  & \multicolumn{13}{c}{BLANK (BLANK; ~\citealt{})} \\ \cmidrule(lr){2-14}
% \multicolumn{1}{c}{\multirow{2}{*}{}} & \multirow{2}{*}{\begin{tabular}[c]{@{}c@{}}Traditional \\ Metric Alone\end{tabular}} & \multicolumn{2}{c}{Agreeableness} & \multicolumn{2}{c}{Empathy} & \multicolumn{2}{c}{Emo. Entropy} & \multicolumn{2}{c}{Emo. Matching} & \multicolumn{2}{c}{Lang. Style Matching} & \multicolumn{2}{c}{All Human} \\ \cmidrule(lr){3-4} \cmidrule(lr){5-6} \cmidrule(lr){7-8} \cmidrule(lr){9-10} \cmidrule(lr){11-12}\cmidrule(lr){13-14}
% \multicolumn{1}{c}{} &  & H & H+A & H & H+A & H & H+A & H & H+A & H & H+A & H & H+A \\ \hline
% BARTScore     & . & . & . & . & . & . & . & . & . & . & . & . & . \\
% BERTScore          & . & . & . & . & . & . & . & . & . & . & . & . & . \\
% BLEURT        & . & . & . & . & . & . & . & . & . & . & . & . & . \\
% DialogRPT & . & . & . & . & . & . & . & . & . & . & . & . & . \\
% FED & . & . & . & . & . & . & . & . & . & . & . & . & . \\
% Prism Ref.    & . & . & . & . & . & . & . & . & . & . & . & . & . \\
% Prism Unref.  & . & . & . & . & . & . & . & . & . & . & . & . & . \\
% Prism Context & . & . & . & . & . & . & . & . & . & . & . & . & . \\ 
% USL-H & . & . & . & . & . & . & . & . & . & . & . & . & . \\
% USR & . & . & . & . & . & . & . & . & . & . & . & . & . \\ \bottomrule
% \end{tabular}
% }
% \caption{BLANK}
% \label{tab:full results daily dialog gupta}
% \end{table*}

\begin{table*}[ht]
\centering
\resizebox{.98\textwidth}{!}{ 
\begin{tabular}{lccccccccccccc}\toprule
 & \multicolumn{13}{c}{DSTC6 (D6; ~\cite{hori2017end})} \\ \cmidrule(lr){2-14}
\multicolumn{1}{c}{\multirow{2}{*}{}} & \multirow{2}{*}{\begin{tabular}[c]{@{}c@{}}Traditional \\ Metric Alone\end{tabular}} & \multicolumn{2}{c}{Agreeableness} & \multicolumn{2}{c}{Empathy} & \multicolumn{2}{c}{Emo. Entropy} & \multicolumn{2}{c}{Emo. Matching} & \multicolumn{2}{c}{Lang. Style Matching} & \multicolumn{2}{c}{All Psych.} \\ \cmidrule(lr){3-4} \cmidrule(lr){5-6} \cmidrule(lr){7-8} \cmidrule(lr){9-10} \cmidrule(lr){11-12}\cmidrule(lr){13-14}
\multicolumn{1}{c}{} &  & P & P+T & P & P+T & P & P+T & P & P+T & P & P+T & P & P+T \\ \hline
BARTScore & .080 & .000 & .080$^{*}$ & .001 & .081$^{***}$ & .038 & .103$^{***}$ & .002 & .081 & .006 & .084$^{***}$ & .009 & .110$^{***}$ \\
BERTScore & .195 & .000 & .196$^{*}$ & .001 & .195 & .038 & .224$^{***}$ & .002 & .196 & .006 & .200$^{***}$ & .009 & .227$^{***}$ \\
BLEURT & .167 & .000 & .167 & .001 & .168$^{***}$ & .038 & .183$^{***}$ & .002 & .168 & .006 & .170$^{***}$ & .009 & .187$^{***}$ \\
Prism ref. & .081 & .000 & .082 & .001 & .084$^{***}$ & .038 & .093$^{***}$ & .002 & .083 & .006 & .085$^{***}$ & .009 & .101$^{***}$ \\
Prism Unref. & .024 & .000 & .024 & .001 & .026$^{***}$ & .038 & .044$^{***}$ & .002 & .026 & .006 & .029$^{***}$ & .009 & .052$^{***}$ \\
Prism Context & .014 & .000 & .015 & .001 & .016$^{***}$ & .038 & .042$^{***}$ & .002 & .016 & .006 & .019$^{***}$ & .009 & .050$^{***}$ \\
\bottomrule
\end{tabular}
}
\caption{DSTC6 data set, reported linear regression adjusted $R^2$ where $P$ contains the psychological metric as the independent variable and $P+T$ contains both the psychological and traditional metrics as independent variables. Benjamini-Hochberg corrected significance level: $^{***}$ p < 0.001, $^{**}$  p < 0.01, $^{*}$ p < 0.05  }
\label{tab:full results DSTC6}
\end{table*}

\begin{table*}[ht]
\centering
\resizebox{.98\textwidth}{!}{ 
\begin{tabular}{lccccccccccccc}\toprule
 & \multicolumn{13}{c}{DSTC7 (D7; ~\cite{galley2019grounded})} \\ \cmidrule(lr){2-14}
\multicolumn{1}{c}{\multirow{2}{*}{}} & \multirow{2}{*}{\begin{tabular}[c]{@{}c@{}}Traditional \\ Metric Alone\end{tabular}} & \multicolumn{2}{c}{Agreeableness} & \multicolumn{2}{c}{Empathy} & \multicolumn{2}{c}{Emo. Entropy} & \multicolumn{2}{c}{Emo. Matching} & \multicolumn{2}{c}{Lang. Style Matching} & \multicolumn{2}{c}{All Psych.} \\ \cmidrule(lr){3-4} \cmidrule(lr){5-6} \cmidrule(lr){7-8} \cmidrule(lr){9-10} \cmidrule(lr){11-12}\cmidrule(lr){13-14}
\multicolumn{1}{c}{} &  & P & P+T & P & P+T & P & P+T & P & P+T & P & P+T & P & P+T \\ \hline
BARTScore & .087 & .000 & .088 & .000 & .087 & .023 & .095$^{***}$ & .001 & .088 & .000 & .091$^{***}$ & .009 & .099$^{***}$ \\
BERTScore & .130 & .000 & .131 & .000 & .130 & .023 & .140$^{***}$ & .001 & .131 & .000 & .130 & .009 & .141$^{***}$ \\
BLEURT & .126 & .000 & .127 & .000 & .126 & .023 & .130$^{**}$ & .001 & .127 & .000 & .126 & .009 & .131$^{***}$ \\
Prism ref. & .101 & .000 & .101 & .000 & .101 & .023 & .105$^{**}$ & .001 & .101 & .000 & .101 & .009 & .106$^{***}$ \\
Prism Unref. & .021 & .000 & .021 & .000 & .021 & .023 & .028$^{***}$ & .001 & .021 & .000 & .021 & .009 & .029$^{***}$ \\
Prism Context & .011 & .000 & .011 & .000 & .011 & .023 & .026$^{***}$ & .001 & .011 & .000 & .012 & .009 & .028$^{***}$ \\
\bottomrule
\end{tabular}
}
\caption{DSTC7 data set, reported linear regression adjusted $R^2$ where $P$ contains the psychological metric as the independent variable and $P+T$ contains both the psychological and traditional metrics as independent variables. Benjamini-Hochberg corrected significance level: $^{***}$ p < 0.001, $^{**}$  p < 0.01, $^{*}$ p < 0.05  }
\label{tab:full results DSTC7}
\end{table*}

\begin{table*}[ht]
\centering
\resizebox{.98\textwidth}{!}{ 
\begin{tabular}{lccccccccccccc}\toprule
 & \multicolumn{13}{c}{English As a Second Language (ESL; ~\cite{zhang2021auteval})} \\ \cmidrule(lr){2-14}
\multicolumn{1}{c}{\multirow{2}{*}{}} & \multirow{2}{*}{\begin{tabular}[c]{@{}c@{}}Traditional \\ Metric Alone\end{tabular}} & \multicolumn{2}{c}{Agreeableness} & \multicolumn{2}{c}{Empathy} & \multicolumn{2}{c}{Emo. Entropy} & \multicolumn{2}{c}{Emo. Matching} & \multicolumn{2}{c}{Lang. Style Matching} & \multicolumn{2}{c}{All Psych.} \\ \cmidrule(lr){3-4} \cmidrule(lr){5-6} \cmidrule(lr){7-8} \cmidrule(lr){9-10} \cmidrule(lr){11-12}\cmidrule(lr){13-14}
\multicolumn{1}{c}{} &  & P & P+T & P & P+T & P & P+T & P & P+T & P & P+T & P & P+T \\ \hline
BARTScore & .182 & .002 & .182 & .005 & .182 & .004 & .182 & .000 & .185 & .011 & .192 & .009 & .191 \\
BERTScore & .096 & .002 & .098 & .005 & .098 & .004 & .103 & .000 & .106 & .011 & .127 & .009 & .131$^{*}$ \\
BLEURT & .080 & .002 & .082$^{*}$ & .005 & .084 & .004 & .081 & .000 & .081 & .011 & .094 & .009 & .098$^{*}$ \\
Prism ref. & .066 & .002 & .067 & .005 & .067 & .004 & .067 & .000 & .067 & .011 & .079 & .009 & .078 \\
Prism Unref. & .011 & .002 & .012 & .005 & .015 & .004 & .010 & .000 & .011 & .011 & .020 & .009 & .023$^{*}$ \\
Prism Context & .007 & .002 & .009 & .005 & .011 & .004 & .013 & .000 & .013 & .011 & .036 & .009 & .040$^{*}$ \\
\bottomrule
\end{tabular}
}
\caption{ESL data set, reported linear regression adjusted $R^2$ where $P$ contains the psychological metric as the independent variable and $P+T$ contains both the psychological and traditional metrics as independent variables. Benjamini-Hochberg corrected significance level: $^{***}$ p < 0.001, $^{**}$  p < 0.01, $^{*}$ p < 0.05 }
\label{tab:full results esl}
\end{table*}

\begin{table*}[ht]
\centering
\resizebox{.98\textwidth}{!}{ 
\begin{tabular}{lccccccccccccc}\toprule
 & \multicolumn{13}{c}{DailyDialog (GD; ~\cite{gupta2019investigating})} \\ \cmidrule(lr){2-14}
\multicolumn{1}{c}{\multirow{2}{*}{}} & \multirow{2}{*}{\begin{tabular}[c]{@{}c@{}}Traditional \\ Metric Alone\end{tabular}} & \multicolumn{2}{c}{Agreeableness} & \multicolumn{2}{c}{Empathy} & \multicolumn{2}{c}{Emo. Entropy} & \multicolumn{2}{c}{Emo. Matching} & \multicolumn{2}{c}{Lang. Style Matching} & \multicolumn{2}{c}{All Psych.} \\ \cmidrule(lr){3-4} \cmidrule(lr){5-6} \cmidrule(lr){7-8} \cmidrule(lr){9-10} \cmidrule(lr){11-12}\cmidrule(lr){13-14}
\multicolumn{1}{c}{} &  & P & P+T & P & P+T & P & P+T & P & P+T & P & P+T & P & P+T \\ \hline
BARTScore & .017 & .005 & .021 & .000 & .018 & .005 & .024$^{*}$ & .002 & .020 & .004 & .025 & .009 & .038$^{**}$ \\
BERTScore & .116 & .005 & .119 & .000 & .116 & .005 & .115 & .002 & .117 & .004 & .119 & .009 & .122$^{*}$ \\
BLEURT & .121 & .005 & .123 & .000 & .121 & .005 & .135$^{*}$ & .002 & .120 & .004 & .129 & .009 & .140$^{**}$ \\
Prism ref. & .014 & .005 & .018 & .000 & .015 & .005 & .022$^{*}$ & .002 & .016 & .004 & .022 & .009 & .033$^{**}$ \\
Prism Unref. & .002 & .005 & .003 & .000 & .002 & .005 & .004$^{*}$ & .002 & .001 & .004 & .002 & .009 & .012$^{**}$ \\
Prism Context & .018 & .005 & .025 & .000 & .018 & .005 & .017 & .002 & .024 & .004 & .018 & .009 & .027$^{**}$ \\ 
\bottomrule
\end{tabular}
}
\caption{DailyDialog (GD) data set, reported linear regression adjusted $R^2$ where $P$ contains the psychological metric as the independent variable and $P+T$ contains both the psychological and traditional metrics as independent variables. Benjamini-Hochberg corrected significance level: $^{***}$ p < 0.001, $^{**}$  p < 0.01, $^{*}$ p < 0.05  }
\label{tab:full results daily dialog gupta}
\end{table*}

\begin{table*}[ht]
\centering
\resizebox{.98\textwidth}{!}{ 
\begin{tabular}{lccccccccccccc}\toprule
 & \multicolumn{13}{c}{HUMOD (HU; ~\cite{merdivan2020human})} \\ \cmidrule(lr){2-14}
\multicolumn{1}{c}{\multirow{2}{*}{}} & \multirow{2}{*}{\begin{tabular}[c]{@{}c@{}}Traditional \\ Metric Alone\end{tabular}} & \multicolumn{2}{c}{Agreeableness} & \multicolumn{2}{c}{Empathy} & \multicolumn{2}{c}{Emo. Entropy} & \multicolumn{2}{c}{Emo. Matching} & \multicolumn{2}{c}{Lang. Style Matching} & \multicolumn{2}{c}{All Psych.} \\ \cmidrule(lr){3-4} \cmidrule(lr){5-6} \cmidrule(lr){7-8} \cmidrule(lr){9-10} \cmidrule(lr){11-12}\cmidrule(lr){13-14}
\multicolumn{1}{c}{} &  & P & P+T & P & P+T & P & P+T & P & P+T & P & P+T & P & P+T \\ \hline
BARTScore & .078 & .000 & .078 & .000 & .078 & .006 & .086$^{***}$ & .013 & .089$^{***}$ & .002 & .078 & .009 & .099$^{***}$ \\
BERTScore & .120 & .000 & .121 & .000 & .120 & .006 & .121$^{*}$ & .013 & .129$^{***}$ & .002 & .122$^{*}$ & .009 & .130$^{***}$ \\
BLEURT & .123 & .000 & .123 & .000 & .123 & .006 & .129$^{***}$ & .013 & .132$^{***}$ & .002 & .124 & .009 & .139$^{***}$ \\
Prism ref. & .062 & .000 & .062 & .000 & .062 & .006 & .075$^{***}$ & .013 & .073$^{***}$ & .002 & .062 & .009 & .088$^{***}$ \\
Prism Unref. & .006 & .000 & .006 & .000 & .006 & .006 & .008$^{**}$ & .013 & .021$^{***}$ & .002 & .009$^{**}$ & .009 & .026$^{***}$ \\
Prism Context & .024 & .000 & .024 & .000 & .024 & .006 & .043$^{***}$ & .013 & .033$^{***}$ & .002 & .024 & .009 & .052$^{***}$ \\
\bottomrule
\end{tabular}
}
\caption{HUMOD (HU) data set, reported linear regression adjusted $R^2$ where $P$ contains the psychological metric as the independent variable and $P+T$ contains both the psychological and traditional metrics as independent variables. Benjamini-Hochberg corrected significance level: $^{***}$ p < 0.001, $^{**}$  p < 0.01, $^{*}$ p < 0.05  }
\label{tab:full results data set 4 humod}
\end{table*}

\begin{table*}[ht]
\centering
\resizebox{.98\textwidth}{!}{ 
\begin{tabular}{lccccccccccccc}\toprule
 & \multicolumn{13}{c}{Neural Conversation Model (NCM; ~\cite{zhang2021auteval})} \\ \cmidrule(lr){2-14}
\multicolumn{1}{c}{\multirow{2}{*}{}} & \multirow{2}{*}{\begin{tabular}[c]{@{}c@{}}Traditional \\ Metric Alone\end{tabular}} & \multicolumn{2}{c}{Agreeableness} & \multicolumn{2}{c}{Empathy} & \multicolumn{2}{c}{Emo. Entropy} & \multicolumn{2}{c}{Emo. Matching} & \multicolumn{2}{c}{Lang. Style Matching} & \multicolumn{2}{c}{All Psych.} \\ \cmidrule(lr){3-4} \cmidrule(lr){5-6} \cmidrule(lr){7-8} \cmidrule(lr){9-10} \cmidrule(lr){11-12}\cmidrule(lr){13-14}
\multicolumn{1}{c}{} &  & P & P+T & P & P+T & P & P+T & P & P+T & P & P+T & P & P+T \\ \hline
BARTScore & .035 & .001 & .035 & .000 & .035 & .009 & .039$^{*}$ & .006 & .041$^{**}$ & .001 & .035 & .009 & .045 \\
BERTScore & .013 & .001 & .014 & .000 & .013 & .009 & .026 & .006 & .019$^{*}$ & .001 & .014 & .009 & .033 \\
BLEURT & .019 & .001 & .019 & .000 & .020 & .009 & .032$^{*}$ & .006 & .026$^{**}$ & .001 & .019 & .009 & .039 \\
Prism ref. & .019 & .001 & .020 & .000 & .020 & .009 & .030$^{*}$ & .006 & .027$^{**}$ & .001 & .019 & .009 & .038 \\
Prism Unref. & .004 & .001 & .005 & .000 & .004 & .009 & .009 & .006 & .009$^{**}$ & .001 & .005 & .009 & .015 \\
Prism Context & .013 & .001 & .014 & .000 & .013 & .009 & .015$^{*}$ & .006 & .015$^{*}$ & .001 & .015 & .009 & .020 \\
\bottomrule
\end{tabular}
}
\caption{Neural Conversation Model (NCM) data set, reported linear regression adjusted $R^2$ where $P$ contains the psychological metric as the independent variable and $P+T$ contains both the psychological and traditional metrics as independent variables. Benjamini-Hochberg corrected significance level: $^{***}$ p < 0.001, $^{**}$  p < 0.01, $^{*}$ p < 0.05 }
\label{tab:full results ncm}
\end{table*}

\begin{table*}[ht]
\centering
\resizebox{.98\textwidth}{!}{ 
\begin{tabular}{lccccccccccccc}\toprule
 & \multicolumn{13}{c}{Persona-DSTC10 (PD10; ~\cite{zhang2021auteval})} \\ \cmidrule(lr){2-14}
\multicolumn{1}{c}{\multirow{2}{*}{}} & \multirow{2}{*}{\begin{tabular}[c]{@{}c@{}}Traditional \\ Metric Alone\end{tabular}} & \multicolumn{2}{c}{Agreeableness} & \multicolumn{2}{c}{Empathy} & \multicolumn{2}{c}{Emo. Entropy} & \multicolumn{2}{c}{Emo. Matching} & \multicolumn{2}{c}{Lang. Style Matching} & \multicolumn{2}{c}{All Psych.} \\ \cmidrule(lr){3-4} \cmidrule(lr){5-6} \cmidrule(lr){7-8} \cmidrule(lr){9-10} \cmidrule(lr){11-12}\cmidrule(lr){13-14}
\multicolumn{1}{c}{} &  & P & P+T & P & P+T & P & P+T & P & P+T & P & P+T & P & P+T \\ \hline
BARTScore & .060 & .000 & .060 & .000 & .060 & .019 & .071$^{**}$ & .012 & .069$^{**}$ & .006 & .062 & .009 & .078$^{***}$ \\
BERTScore & .025 & .000 & .025 & .000 & .025 & .019 & .041$^{**}$ & .012 & .037 & .006 & .031 & .009 & .050$^{**}$ \\
BLEURT & .043 & .000 & .043 & .000 & .043 & .019 & .054$^{***}$ & .012 & .053$^{**}$ & .006 & .048 & .009 & .062$^{***}$ \\
Prism ref. & .019 & .000 & .019 & .000 & .019 & .019 & .035$^{**}$ & .012 & .031 & .006 & .024 & .009 & .044$^{**}$ \\
Prism Unref. & .020 & .000 & .020 & .000 & .020 & .019 & .028$^{**}$ & .012 & .030$^{*}$ & .006 & .024 & .009 & .037$^{**}$ \\
Prism Context & .008 & .000 & .008 & .000 & .008 & .019 & .027$^{**}$ & .012 & .018 & .006 & .012 & .009 & .033$^{**}$ \\
\bottomrule
\end{tabular}
}
\caption{Persona-DSTC10, reported linear regression adjusted $R^2$ where $P$ contains the psychological metric as the independent variable and $P+T$ contains both the psychological and traditional metrics as independent variables. Benjamini-Hochberg corrected significance level: $^{***}$ p < 0.001, $^{**}$  p < 0.01, $^{*}$ p < 0.05 }
\label{tab:full results persona}
\end{table*}

\begin{table*}[ht]
\centering
\resizebox{.98\textwidth}{!}{ 
\begin{tabular}{lccccccccccccc}\toprule
 & \multicolumn{13}{c}{Topical-DSTC10 (TD10; ~\cite{zhang2021auteval})} \\ \cmidrule(lr){2-14}
\multicolumn{1}{c}{\multirow{2}{*}{}} & \multirow{2}{*}{\begin{tabular}[c]{@{}c@{}}Traditional \\ Metric Alone\end{tabular}} & \multicolumn{2}{c}{Agreeableness} & \multicolumn{2}{c}{Empathy} & \multicolumn{2}{c}{Emo. Entropy} & \multicolumn{2}{c}{Emo. Matching} & \multicolumn{2}{c}{Lang. Style Matching} & \multicolumn{2}{c}{All Psych.} \\ \cmidrule(lr){3-4} \cmidrule(lr){5-6} \cmidrule(lr){7-8} \cmidrule(lr){9-10} \cmidrule(lr){11-12}\cmidrule(lr){13-14}
\multicolumn{1}{c}{} &  & P & P+T & P & P+T & P & P+T & P & P+T & P & P+T & P & P+T \\ \hline
BARTScore & .063 & .000 & .063 & .000 & .063 & .000 & .063 & .001 & .063 & .002 & .063 & .009 & .062 \\
BERTScore & .054 & .000 & .054 & .000 & .054 & .000 & .054 & .001 & .055 & .002 & .055 & .009 & .055 \\
BLEURT & .049 & .000 & .049 & .000 & .049 & .000 & .049 & .001 & .049 & .002 & .049 & .009 & .049 \\
Prism ref. & .036 & .000 & .036 & .000 & .036 & .000 & .036 & .001 & .037 & .002 & .037 & .009 & .037 \\
Prism Unref. & .000 & .000 & .000 & .000 & .000 & .000 & .000 & .001 & .000 & .002 & .002 & .009 & .002 \\
Prism Context & .005 & .000 & .005 & .000 & .005 & .000 & .005 & .001 & .005 & .002 & .006 & .009 & .005 \\
 \bottomrule
\end{tabular}
}
\caption{Topical-DSTC10, reported linear regression adjusted $R^2$ where $P$ contains the psychological metric as the independent variable and $P+T$ contains both the psychological and traditional metrics as independent variables. Benjamini-Hochberg corrected significance level: $^{***}$ p < 0.001, $^{**}$  p < 0.01, $^{*}$ p < 0.05 }
\label{tab:full results topical}
\end{table*}

\begin{table*}[ht]
\centering
\resizebox{.98\textwidth}{!}{ 
\begin{tabular}{lccccccccccccc}\toprule
 & \multicolumn{13}{c}{TopicalChat-USR (TP; ~\cite{mehri2020usr})} \\ \cmidrule(lr){2-14}
\multicolumn{1}{c}{\multirow{2}{*}{}} & \multirow{2}{*}{\begin{tabular}[c]{@{}c@{}}Traditional \\ Metric Alone\end{tabular}} & \multicolumn{2}{c}{Agreeableness} & \multicolumn{2}{c}{Empathy} & \multicolumn{2}{c}{Emo. Entropy} & \multicolumn{2}{c}{Emo. Matching} & \multicolumn{2}{c}{Lang. Style Matching} & \multicolumn{2}{c}{All Psych.} \\ \cmidrule(lr){3-4} \cmidrule(lr){5-6} \cmidrule(lr){7-8} \cmidrule(lr){9-10} \cmidrule(lr){11-12}\cmidrule(lr){13-14}
\multicolumn{1}{c}{} &  & P & P+T & P & P+T & P & P+T & P & P+T & P & P+T & P & P+T \\ \hline
BARTScore & .138 & .017 & .146 & .000 & .136 & .110 & .213$^{*}$ & .003 & .136 & .070 & .168$^{*}$ & .009 & .227$^{**}$ \\
BERTScore & .217 & .017 & .224 & .000 & .215 & .110 & .279$^{*}$ & .003 & .215 & .070 & .247$^{*}$ & .009 & .295$^{**}$ \\
BLEURT & .283 & .017 & .289 & .000 & .281 & .110 & .324$^{*}$ & .003 & .282 & .070 & .298$^{*}$ & .009 & .332$^{**}$ \\
Prism ref. & .177 & .017 & .186 & .000 & .175 & .110 & .236$^{*}$ & .003 & .175 & .070 & .204$^{*}$ & .009 & .252$^{**}$ \\
Prism Unref. & .204 & .017 & .209 & .000 & .202 & .110 & .238$^{*}$ & .003 & .202 & .070 & .229 & .009 & .252 \\
Prism Context & .018 & .017 & .033 & .000 & .017 & .110 & .117$^{**}$ & .003 & .015 & .070 & .099$^{**}$ & .009 & .162$^{***}$ \\
 \bottomrule
\end{tabular}
}
\caption{TopicalChat data set, reported linear regression adjusted $R^2$ where $P$ contains the psychological metric as the independent variable and $P+T$ contains both the psychological and traditional metrics as independent variables. Benjamini-Hochberg corrected significance level: $^{***}$ p < 0.001, $^{**}$  p < 0.01, $^{*}$ p < 0.05 }
\label{tab:full results topicalchat}
\end{table*}

\begin{table*}[ht]
\centering
\resizebox{.98\textwidth}{!}{ 
\begin{tabular}{lccccccccccccc}\toprule
 & \multicolumn{13}{c}{PersonaChat-USR (UP; ~\cite{mehri2020usr})} \\ \cmidrule(lr){2-14}
\multicolumn{1}{c}{\multirow{2}{*}{}} & \multirow{2}{*}{\begin{tabular}[c]{@{}c@{}}Traditional \\ Metric Alone\end{tabular}} & \multicolumn{2}{c}{Agreeableness} & \multicolumn{2}{c}{Empathy} & \multicolumn{2}{c}{Emo. Entropy} & \multicolumn{2}{c}{Emo. Matching} & \multicolumn{2}{c}{Lang. Style Matching} & \multicolumn{2}{c}{All Psych.} \\ \cmidrule(lr){3-4} \cmidrule(lr){5-6} \cmidrule(lr){7-8} \cmidrule(lr){9-10} \cmidrule(lr){11-12}\cmidrule(lr){13-14}
\multicolumn{1}{c}{} &  & P & P+T & P & P+T & P & P+T & P & P+T & P & P+T & P & P+T \\ \hline
BARTScore & .044 & .003 & .041 & .003 & .041 & .130 & .179$^{**}$ & .003 & .042 & .019 & .060 & .009 & .183$^{***}$ \\
BERTScore & .109 & .003 & .107 & .003 & .106 & .130 & .230$^{**}$ & .003 & .107 & .019 & .128 & .009 & .238$^{***}$ \\
BLEURT & .151 & .003 & .150 & .003 & .149 & .130 & .258$^{**}$ & .003 & .148 & .019 & .161 & .009 & .260$^{***}$ \\
Prism ref. & .080 & .003 & .077 & .003 & .077 & .130 & .202$^{**}$ & .003 & .077 & .019 & .097 & .009 & .208$^{***}$ \\
Prism Unref. & .098 & .003 & .095 & .003 & .097 & .130 & .159$^{*}$ & .003 & .096 & .019 & .116$^{*}$ & .009 & .168$^{***}$ \\
Prism Context & .023 & .003 & .020 & .003 & .020 & .130 & .141$^{**}$ & .003 & .022 & .019 & .054$^{*}$ & .009 & .157$^{***}$ \\  \bottomrule
\end{tabular}
}
\caption{PersonaChat-USR (UP) data set, reported linear regression adjusted $R^2$ where $P$ contains the psychological metric as the independent variable and $P+T$ contains both the psychological and traditional metrics as independent variables. Benjamini-Hochberg corrected significance level: $^{***}$ p < 0.001, $^{**}$  p < 0.01, $^{*}$ p < 0.05 }
\label{tab:full results personachat usr}
\end{table*}

\begin{table*}[ht]
\centering
\resizebox{.98\textwidth}{!}{ 
\begin{tabular}{lccccccccccccc}\toprule
 & \multicolumn{13}{c}{DailyDialog (ZD; ~\cite{zhao2020designing})} \\ \cmidrule(lr){2-14}
\multicolumn{1}{c}{\multirow{2}{*}{}} & \multirow{2}{*}{\begin{tabular}[c]{@{}c@{}}Traditional \\ Metric Alone\end{tabular}} & \multicolumn{2}{c}{Agreeableness} & \multicolumn{2}{c}{Empathy} & \multicolumn{2}{c}{Emo. Entropy} & \multicolumn{2}{c}{Emo. Matching} & \multicolumn{2}{c}{Lang. Style Matching} & \multicolumn{2}{c}{All Psych.} \\ \cmidrule(lr){3-4} \cmidrule(lr){5-6} \cmidrule(lr){7-8} \cmidrule(lr){9-10} \cmidrule(lr){11-12}\cmidrule(lr){13-14}
\multicolumn{1}{c}{} &  & P & P+T & P & P+T & P & P+T & P & P+T & P & P+T & P & P+T \\ \hline
BARTScore & .148 & .001 & .149 & .000 & .147 & .020 & .192$^{***}$ & .004 & .150 & .013 & .172$^{**}$ & .009 & .204$^{***}$ \\
BERTScore & .171 & .001 & .171 & .000 & .170 & .020 & .196$^{**}$ & .004 & .172 & .013 & .191$^{**}$ & .009 & .208$^{***}$ \\
BLEURT & .208 & .001 & .207 & .000 & .207 & .020 & .245$^{***}$ & .004 & .208 & .013 & .235$^{**}$ & .009 & .258$^{***}$ \\
Prism ref. & .146 & .001 & .145 & .000 & .145 & .020 & .184$^{***}$ & .004 & .146 & .013 & .169$^{**}$ & .009 & .194$^{***}$ \\
Prism Unref. & .024 & .001 & .025 & .000 & .024 & .020 & .029$^{**}$ & .004 & .029 & .013 & .031$^{*}$ & .009 & .041$^{***}$ \\
Prism Context & .007 & .001 & .008 & .000 & .007 & .020 & .039$^{***}$ & .004 & .010 & .013 & .024$^{**}$ & .009 & .051$^{***}$ \\
  \bottomrule
\end{tabular}
}
\caption{DailyDialog (ZD) data set, reported linear regression adjusted $R^2$ where $P$ contains the psychological metric as the independent variable and $P+T$ contains both the psychological and traditional metrics as independent variables. Benjamini-Hochberg corrected significance level: $^{***}$ p < 0.001, $^{**}$  p < 0.01, $^{*}$ p < 0.05 }
\label{tab:full results daily dialog zhao}
\end{table*}

\begin{table*}[ht]
\centering
\resizebox{.98\textwidth}{!}{ 
\begin{tabular}{lccccccccccccc}\toprule
 & \multicolumn{13}{c}{PersonaChat (ZP; ~\cite{zhao2020designing})} \\ \cmidrule(lr){2-14}
\multicolumn{1}{c}{\multirow{2}{*}{}} & \multirow{2}{*}{\begin{tabular}[c]{@{}c@{}}Traditional \\ Metric Alone\end{tabular}} & \multicolumn{2}{c}{Agreeableness} & \multicolumn{2}{c}{Empathy} & \multicolumn{2}{c}{Emo. Entropy} & \multicolumn{2}{c}{Emo. Matching} & \multicolumn{2}{c}{Lang. Style Matching} & \multicolumn{2}{c}{All Psych.} \\ \cmidrule(lr){3-4} \cmidrule(lr){5-6} \cmidrule(lr){7-8} \cmidrule(lr){9-10} \cmidrule(lr){11-12}\cmidrule(lr){13-14}
\multicolumn{1}{c}{} &  & P & P+T & P & P+T & P & P+T & P & P+T & P & P+T & P & P+T \\ \hline
BARTScore & .179 & .002 & .182 & .001 & .178 & .000 & .180 & .025 & .202$^{***}$ & .040 & .200$^{**}$ & .009 & .224$^{***}$ \\
BERTScore & .160 & .002 & .163 & .001 & .159 & .000 & .163 & .025 & .184$^{***}$ & .040 & .187$^{**}$ & .009 & .214$^{***}$ \\
BLEURT & .170 & .002 & .172 & .001 & .169 & .000 & .174 & .025 & .191$^{**}$ & .040 & .191$^{**}$ & .009 & .216$^{***}$ \\
Prism ref. & .132 & .002 & .134 & .001 & .131 & .000 & .134 & .025 & .157$^{***}$ & .040 & .158$^{**}$ & .009 & .184$^{***}$ \\
Prism Unref. & .002 & .002 & .004 & .001 & .001 & .000 & .001 & .025 & .027$^{***}$ & .040 & .040$^{***}$ & .009 & .060$^{***}$ \\
Prism Context & .026 & .002 & .027 & .001 & .025 & .000 & .025 & .025 & .042$^{**}$ & .040 & .058$^{***}$ & .009 & .072$^{***}$ \\
  \bottomrule
\end{tabular}
}
\caption{PersonaChat-ZP (ZP) data set, reported linear regression adjusted $R^2$ where $P$ contains the psychological metric as the independent variable and $P+T$ contains both the psychological and traditional metrics as independent variables. Benjamini-Hochberg corrected significance level: $^{***}$ p < 0.001, $^{**}$  p < 0.01, $^{*}$ p < 0.05 }
\label{tab:full results personachat zp}
\end{table*}

\end{document}